\documentclass[11pt,a4paper]{article}
\usepackage[dvipsnames]{xcolor}
\usepackage[hyperref]{acl2019}

\usepackage{times}
\usepackage{latexsym}
\usepackage{graphicx}
\usepackage{todonotes}
\usepackage{microtype}
\usepackage{url}
\usepackage{amsmath}
\usepackage{algorithm}
\usepackage[noend]{algpseudocode}
\usepackage{float}
\usepackage{enumitem}
\usepackage{booktabs}
\usepackage{subfigure}
\usepackage{xspace}
\usepackage[latin1]{inputenc}   
\usepackage[T1]{fontenc}
\usepackage{multirow}
\usepackage{multicol}
\aclfinalcopy 
\def\aclpaperid{2305} 

\DeclareMathAlphabet\mathbfcal{OMS}{cmsy}{b}{n}

\newcommand{\ent}[1]{\texttt{#1}}
\newcommand{\rel}[1]{{\em\texttt{#1}}}
\newcommand{\tok}[1]{``\emph{#1}''}
\newcommand{\entity}[2]{{\color[HTML]{#2}[\emph{#1}]}}
\newcommand{\ds}{\emph{Linked WikiText-2}\xspace}

\newcommand{\camready}[1]{}
\newcommand{\para}[1]{\vskip 1.5mm\noindent\textbf{#1} }
\newcommand{\cut}[1]{}

\algdef{SE}[SUBALG]{Indent}{EndIndent}{}{\algorithmicend\ }%
\algtext*{Indent}
\algtext*{EndIndent}

\title{Barack's Wife Hillary:\\Using Knowledge Graphs for Fact-Aware Language Modeling}

\def\authorGap{10mm}
\author{
  Robert L. Logan IV\footnotemark[1]
  \hspace{\authorGap}
  Nelson F. Liu\footnotemark[2] \footnotemark[4]
  \hspace{\authorGap}
  Matthew E. Peters\footnotemark[4]\\
  \textbf{
  Matt Gardner\footnotemark[4]
  \hspace{\authorGap}
  Sameer Singh\footnotemark[1]}
  \vspace{ 2mm}
\\
  \footnotemark[1] ~\textnormal{University of California, Irvine, CA, USA} \\
  \footnotemark[2] ~University of Washington, Seattle, WA, USA\\
  \footnotemark[4] ~Allen Institute for Artificial Intelligence, Seattle, WA, USA 
    \vspace{ 1mm}
  \\
  {\hypersetup{urlcolor=black}
  \texttt{\{\href{mailto:rlogan@uci.edu}{rlogan}, \href{mailto:sameer@uci.edu}{sameer}\}@uci.edu},
  \texttt{\{\href{mailto:mattg@allenai.org}{mattg}, \href{mailto:matthewp@allenai.org}{matthewp}\}@allenai.org},
  {\tt \href{mailto:nfliu@cs.washington.edu}{nfliu}@cs.washington.edu}
  }
}

\begin{document}
\maketitle

\begin{abstract}
    Modeling human language requires the ability to not only generate fluent text but also encode factual knowledge.
    However, traditional language models are only capable of remembering facts seen at training time, and often have difficulty recalling them.
    To address this, we introduce the knowledge graph language model (KGLM), a neural language model with mechanisms for selecting and copying facts from a knowledge graph that are relevant to the context.
    These mechanisms enable the model to render information it has never seen before, as well as generate out-of-vocabulary tokens.
    We also introduce the \ds{} dataset,\footnote{\url{https://rloganiv.github.io/linked-wikitext-2}}
    a corpus of annotated text aligned to the Wikidata knowledge graph whose contents (roughly) match the popular \emph{WikiText-2} benchmark~\cite{merity2016pointer}.
    In experiments, we demonstrate that the KGLM achieves significantly better performance than a strong baseline language model.
    We additionally compare different language models' ability to complete sentences requiring factual knowledge, and show that the KGLM outperforms even very large language models in generating facts.
\end{abstract}










\section{Introduction}

\begin{figure}[t!]
    \centering
    \begin{minipage}{0.45\textwidth}
        \entity{Super Mario Land}{1b9e77} is a \entity{1989}{d95f02} \entity{side-scrolling}{7570b3} \entity{platform video game}{e7298a}
        developed and published by \entity{Nintendo}{e6ab02} as a \entity{launch title}{66a61e} for their \entity{Game Boy}{a6761d} \entity{handheld game console}{666666}. 
    \end{minipage}
    \begin{minipage}{0.45\textwidth}
        \centering
        \includegraphics[width=\textwidth]{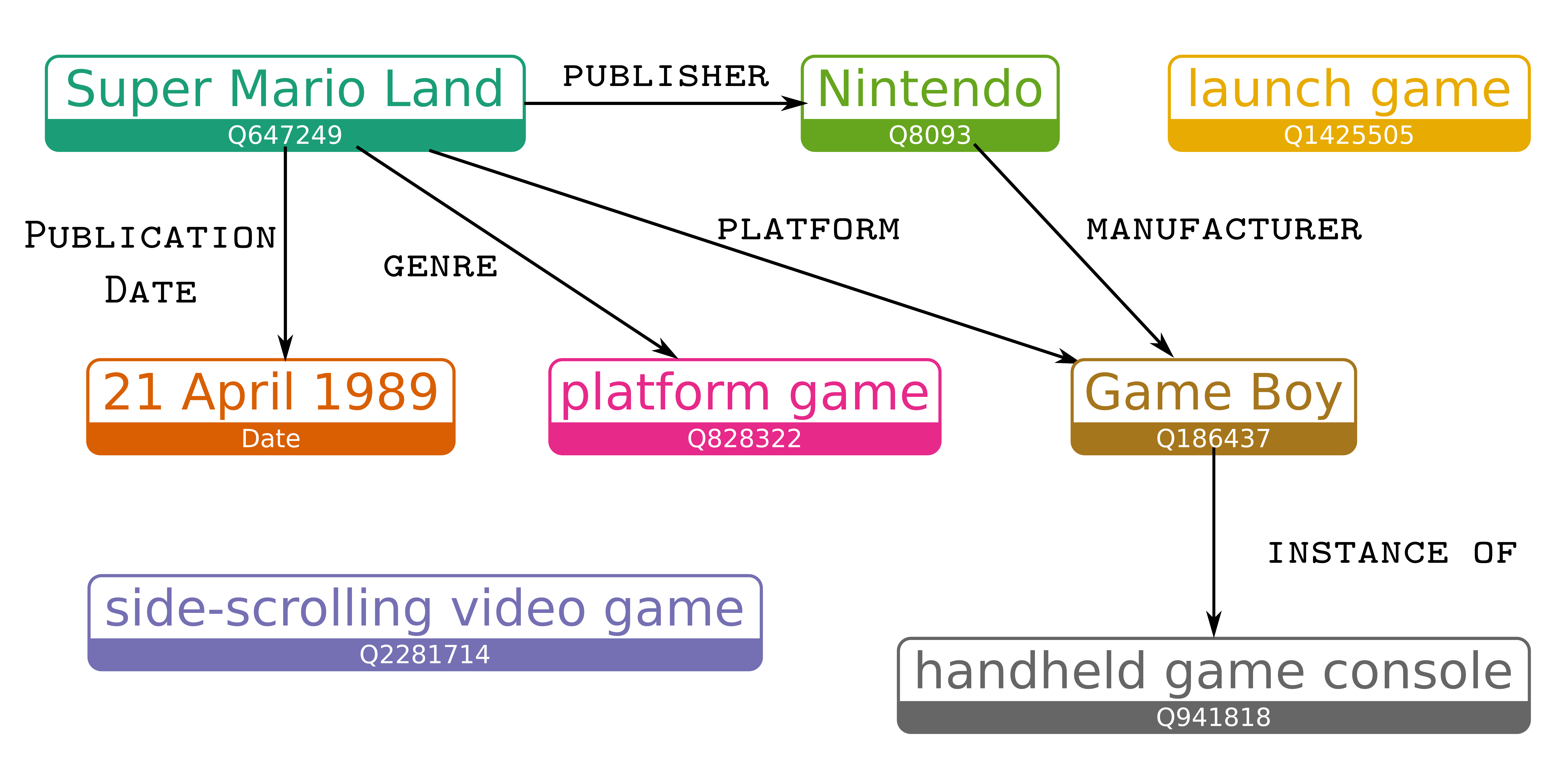}
    \end{minipage}
    \caption{
    {\bf ~\ds{} Example}.
    A localized knowledge graph containing facts that are (possibly) conveyed in the sentence above.
    The graph is built by iteratively linking each detected entity to Wikidata, then adding any relations to previously mentioned entities.
    Note that not all entities are connected, potentially due to missing relations in Wikidata.
    }
    \label{fig:data}
\end{figure}

For language models to generate plausible sentences, they must be both syntactically coherent as well as consistent with the world they describe.
Although language models are quite skilled at generating grammatical sentences, and previous work has shown that language models also possess some degree of common-sense reasoning and basic knowledge~\cite{vinyals2015neural,serban2016building,trinh2019do}, their ability to generate \emph{factually correct} text is quite limited.
The clearest limitation of existing language models is that they, at best, can only memorize facts observed during training.
For instance, when conditioned on the text at the top of Figure~\ref{fig:data}, an AWD-LSTM language model~\cite{merity2017regularizing} trained on \textit{Wikitext-2} assigns higher probability to the word \tok{PlayStation} than \tok{Game Boy}, even though this sentence appears verbatim in the training data.
This is not surprising---existing models represent the distribution over the entire vocabulary directly, whether they are common words, references to real world entities, or factual information like dates and numbers.
As a result, language models are unable to generate factually correct sentences, do not generalize to rare/unseen entities, and often omit rare tokens from the vocabulary (instead generating \emph{UNKNOWN} tokens). 



We introduce the \emph{knowledge graph language model} (KGLM), a neural language model with mechanisms for selecting and copying information from an external knowledge graph.
The KGLM maintains a dynamically growing \emph{local knowledge graph}, a subset of the knowledge graph that contains entities that have already been mentioned in the text, and their related entities.
When generating entity tokens, the model either decides to render a new entity that is absent from the local graph, thereby growing the local knowledge graph,
or to render a fact from the local graph.
When rendering, the model combines the standard vocabulary with tokens available in the knowledge graph, thus supporting numbers, dates, and other rare tokens.

Figure~\ref{fig:data} illustrates how the KGLM works.
Initially, the graph is empty and the model uses the entity \ent{Super Mario Land} to render the first three tokens, thus adding it and its relations to the local knowledge graph.
After generating the next two tokens (\tok{is}, \tok{a}) using the standard language model, the model selects \ent{Super Mario Land} as the parent entity, \rel{Publication Date} as the relation to render, and copies one of the tokens of the date entity as the token (\tok{1989} in this case).

To facilitate research on knowledge graph-based language modeling, we collect the distantly supervised \ds dataset.
The underlying text closely matches \textit{WikiText-2}~\cite{merity2016pointer}, a popular benchmark for language modeling, allowing comparisons against existing models.
The tokens in the text are linked to entities in Wikidata~\cite{wikidata} using a combination of human-provided links and off-the-shelf linking and coreference models. 
We also use relations between these entities in Wikidata to construct plausible reasons for why an entity may have been mentioned:
it could either be related to an entity that is already mentioned (including itself)
or a brand new, unrelated entity for the document.

We train and evaluate the KGLM on \ds.
When compared against AWD-LSTM, a recent and performant language model, KGLM obtains not only a lower overall perplexity, but also a substantially lower \emph{unknown-penalized} perplexity~\cite{UEBERLA1994153,ahn2016neural}, a metric that allows fair comparisons between models that accurately model rare tokens and ones that predict them to be \emph{unknown}.
We also compare \emph{factual completion} capabilities of these models, where they predict the next word after a factual sentence (e.g., \tok{Barack is married to \underline{\hspace{5mm}}}) and show that KGLM is significantly more accurate.
Lastly, we show 
that the model is able to generate accurate facts for rare entities, and can be \emph{controlled} via modifications the knowledge graph.

\begin{figure*}[tb]
    \centering
    \includegraphics[width=0.95\textwidth,clip,trim=40 110 50 120]{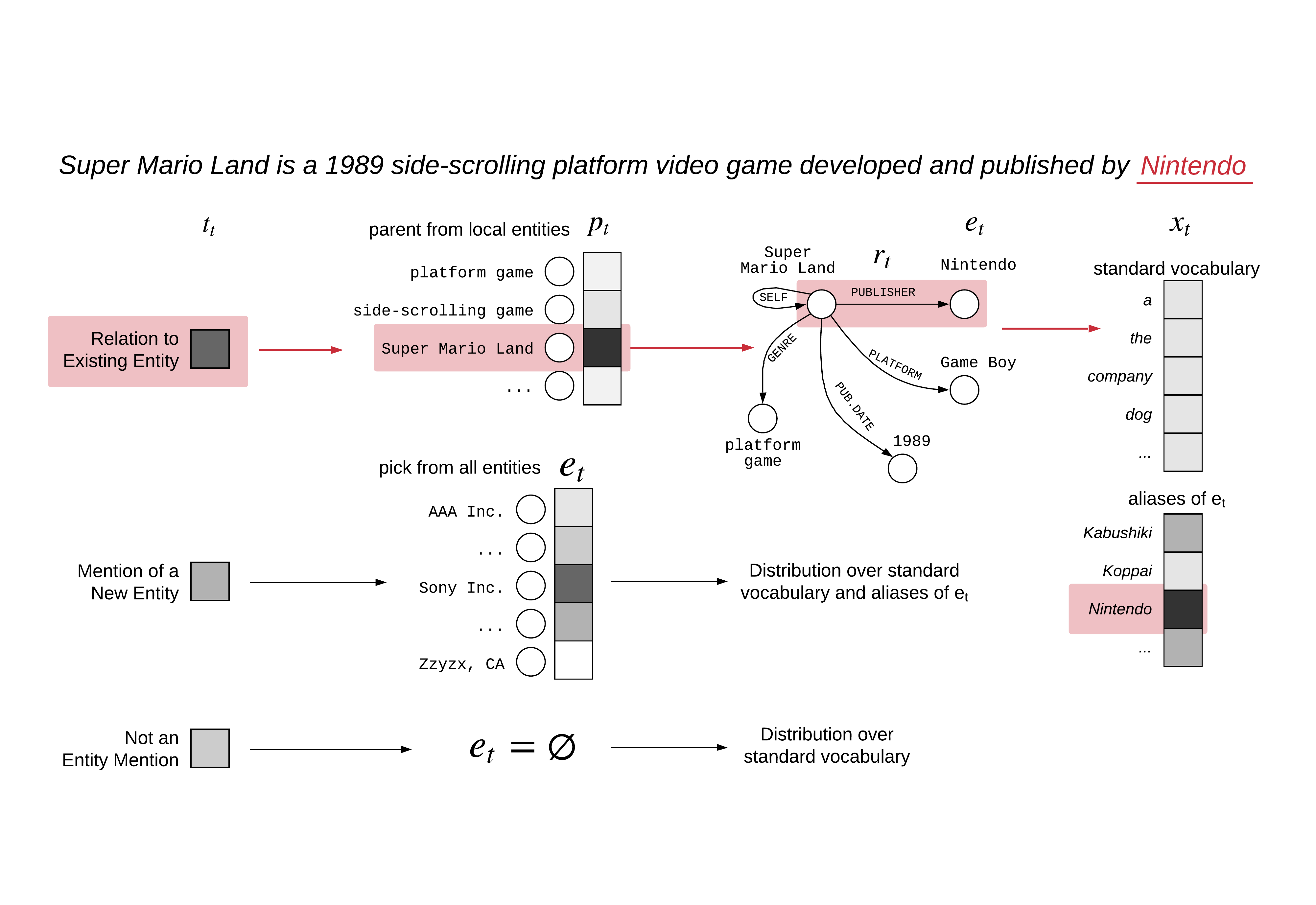}
    \caption{\textbf{KGLM Illustration.} When trying to generate the token following \tok{published by}, the model first decides the type of the mention ($t_t$) to be a related entity (darker indicates higher probability), followed by identifying the parent ($p_t$), relation ($r_t$), and entity to render ($e_t$) from the local knowledge graph as (\ent{Super Mario Land}, \rel{Publisher}, \ent{Nintendo}). The final distribution over the words includes the standard vocabulary along with aliases of \ent{Nintendo}, and the model selects \tok{Nintendo} as the token $x_t$. Facts related to \ent{Nintendo} will be added to the local graph.
    }
    \label{fig:generative}
\end{figure*}

\section{Knowledge Graph Language Model}
\label{sec:model}

In this section we introduce a language model that is conditioned on an external, structured knowledge source, which it uses to generate factual text.

\subsection{Problem Setup and Notation}

A \textit{language model} defines a probability distribution over each token within a sequence, conditioned on the sequence of tokens observed so far.
We denote the random variable representing the next token as $x_t$ and the sequence of the tokens before $t$ as $x_{<t}$, i.e. language models compute $p(x_t|x_{<t})$.
RNN language models~\cite{mikolov2010recurrent} parameterize this distribution using a recurrent structure:
\begin{align}
\label{eqn:rnn-lm}
\begin{split}
    p(x_t | x_{<t}) &= \text{softmax}(\mathbf{W}_h \mathbf{h}_{t} + \mathbf{b}) , 
\\
    \mathbf{h}_{t} &= \text{RNN}(\mathbf{h}_{t-1}, \mathbf{x}_{t-1}) .
\end{split}
\end{align}
We use LSTMs~\cite{hochreiter1997long} as the recurrent module in this paper. 

A \textit{knowledge graph} (KG) is a directed, labeled graph consisting of entities $\mathcal{E}$ as nodes, with edges defined over a set of relations $\mathcal{R}$, i.e. $\mathcal{KG} = \left\{(p, r, e)~|~p \in \mathcal{E}, r \in \mathcal{R}, e \in \mathcal{E} \right\}$, where $p$ is a parent entity with relation $r$ to another entity $e$.
Practical KGs have other aspects that make this formulation somewhat inexact: some relations are to \emph{literal values}, such as numbers and dates, facts may be expressed as \emph{properties} on relations, and entities have \emph{aliases} as the set of strings that can refer to the entity.
We also define a \emph{local knowledge graph} for a subset of entities $\mathcal{E}_{<t}$ as $\mathcal{KG}_{<t} = \left\{(p, r, e)~|~p \in \mathcal{E}_{<t}, r \in \mathcal{R}, e \in \mathcal{E} \right\}$, i.e. contains entities $\mathcal{E}_{<t}$ and all facts they participate in.

\subsection{Generative KG Language Model}
\label{sec:gen-kg}

The primary goal of the knowledge graph language model (KGLM) is to enable a neural language model to generate entities and facts from a knowledge graph. To encourage the model to generate facts that have appeared in the context already, KGLM will maintain a local knowledge graph containing all facts involving entities that have appeared in the context.
As the model decides to refer to entities that have not been referred to yet, it will grow the local knowledge graph with additional entities and facts to reflect the new entity.

Formally, we will compute $p(x_t,\mathcal{E}_t|x_{<t},\mathcal{E}_{<t})$ where $x_{<t}$ is the sequence of observed tokens, $\mathcal{E}_{<t}$ is the set of entities mentioned in $x_{<t}$, and $\mathcal{KG}_{<t}$ is the local knowledge graph determined by $\mathcal{E}_{<t}$, as described above.
The generative process is: 
\begin{itemize}[nosep,leftmargin=5mm]
    \item Decide the \emph{type} of $x_t$, which we denote by $t_t$: whether it is a reference to an entity in $\mathcal{KG}_{<t}$~(\texttt{related}), a reference to an entity not in $\mathcal{KG}_{<t}$~(\texttt{new}), or not an entity mention~($\emptyset$).
    \item If $t_t = \texttt{new}$ then choose the upcoming entity $e_t$ from the set of all entities $\mathcal{E}$.
    \item If $t_t = \texttt{related}$ then:
        \begin{itemize}[nosep]
            \item Choose a parent entity $p_t$ from $\mathcal{E}_{<t}$. 
            \item Choose a factual relation $r_t$ to render,\\
            $r_t\in \left\{ (p, r, e) \in \mathcal{KG}_{<t} | p = p_t \right\}$.
            \item Choose $e_t$ as one of the tail entities,\\
            $e_t\in \left\{ e | (p_t,r_t,e) \in \mathcal{KG}_{<t}\right\}$.
        \end{itemize}
    \item If $t_t = \emptyset$ then $e_t = \emptyset$.
    \item Generate $x_t$ conditioned on $e_t$, potentially copying one of $e_t$'s aliases.
    \item If $e_t\notin\mathcal{E}_{<t}$, then 
        $\mathcal{E}_{<(t+1)}\gets\mathcal{E}_{<t}\cup \left\{ e_t \right\}$, \\
        else $\mathcal{E}_{<(t+1)}\gets\mathcal{E}_{<t}$.
\end{itemize}
For the model to refer to an entity it has already mentioned, we introduce a \rel{Reflexive} relation that self-relates, i.e. $p=e$ for $(p,\text{\rel{Reflexive}},e)$.

An illustration of this process and the variables is provided in Figure~\ref{fig:generative}, for generating a token in the middle of the same sentence as in Figure~\ref{fig:data}.
Amongst the three mention types ($t_t$), the model chooses a reference to existing entity, which requires picking a fact to render.
As the parent entity of this fact ($p_t$), the model picks \ent{Super Mario Land}, and then follows the \rel{Publisher} relation ($r_t$) to select \ent{Nintendo} as the entity to render ($e_t$).
When rendering \ent{Nintendo} as a token $x_t$, the model has an \emph{expanded} vocabulary available to it, containing the standard vocabulary along with all word types in any of the aliases of $e_t$.

\para{Marginalizing out the KG}
There is a mismatch between our initial task requirement, $p(x_t|x_{<t})$, and the model we describe so far, which computes $p(x_t,\mathcal{E}_{t}|x_{<t},\mathcal{E}_{<t})$.
We will essentially \emph{marginalize} out the local knowledge graph to compute the probability of the tokens, i.e. $p(\mathbf{x})=\sum_{\mathbfcal{E}}p(\mathbf{x},\mathbfcal{E})$.
We will clarify this, along with describing the training and the inference/decoding algorithms for this model and other details of the setup, in Section~\ref{sec:algs}.

\subsection{Parameterizing the Distributions}

The parametric distributions used in the generative process above are defined as follows.
We begin by computing the hidden state $\mathbf{h}_t$ using the formula in Eqn~\eqref{eqn:rnn-lm}.
We then split the vector into three components:
$\mathbf{h}_t = \left[\mathbf{h}_{t, x}; \mathbf{h}_{t, p}; \mathbf{h}_{t, r} \right]$,
which are respectively used to predict words, parents, and relations.
The type of the token, $t_t$, is computed using a single-layer softmax over $\mathbf{h}_{t,x}$ to predict one of $\left\{ \texttt{new}, \texttt{related}, \emptyset \right\}$.

\para{Picking an Entity}
We also introduce pretrained embeddings for all entities and relations in the knowledge graph, denoted by $\mathbf{v}_e$ for entity $e$ and $\mathbf{v}_r$ for relation $r$.
To select $e_t$ from all entities in case $t_t=\text{\texttt{new}}$, we use:
$$ p(e_t) = \text{softmax}(\mathbf{v}_e \cdot (\mathbf{h}_{t,p} + \mathbf{h}_{t,r})) $$
over all $e\in\mathcal{E}$.
The reason we add $\mathbf{h}_{t, p}$ and $\mathbf{h}_{t, r}$ is to mimic the structure of TransE, which we use to obtain entity and relation embeddings. 
Details on TransE will be provided in Section~\ref{sec:algs}.
For mention of a related entity, $t_t=\text{\texttt{related}}$, we pick a parent entity $p_t$ using
$$ p(p_t) = \text{softmax}(\mathbf{v}_p\cdot \mathbf{h}_{t,p}) $$
over all $p\in\mathcal{E}_t$, then pick the relation $r_t$ using
$$ p(r_t) = \text{softmax}(\mathbf{v}_r\cdot \mathbf{h}_{t,r}) $$
over all $r\in\{r|(p_t,r,e)\in\mathcal{KG}_t\}$.
The combination of $p_t$ and $r_t$ determine the entity $e_t$ (which must satisfy $(p_t,r_t,e_t)\in\mathcal{KG}_t$; if there are multiple options one is chosen at random).

\para{Rendering the Entity}
If $e_t=\emptyset$, i.e. there is no entity to render, we use the same distribution over the vocabulary as in Eqn~\eqref{eqn:rnn-lm} - a softmax using $\mathbf{h}_{t,x}$.
If there is an entity to render, we construct the distribution over the original vocabulary \emph{and} a vocabulary containing all the tokens that appear in aliases of $e_t$.
This distribution is conditioned on $e_t$ in addition to $x_t$.
To compute the scores over the original vocabulary, $\mathbf{h}_{t,x}$ is replaced by $\mathbf{h}'_{t,x} = \mathbf{W}_{\text{proj}}[\mathbf{h}_{t,x}; \mathbf{v}_{e_t}]$ where $\mathbf{W}_{\text{proj}}$ is a learned weight matrix that projects the concatenated vector into the same vector space as $\mathbf{h}_{t,x}$.

To obtain probabilities for words in the alias vocabulary, we use a copy mechanism~\citet{gu2016incorporating}.
The token sequences comprising each alias $\left\{a_j\right\}$ are embedded then encoded using an LSTM to form vectors $\mathbf{a}_j$.
Copy scores are computed as:
$$ p(x_t = a_j) \propto \exp \left[ \sigma \left( \left( \mathbf{h}'_{t,x} \right)^{T} \mathbf{W}_{\text{copy}} \right) \mathbf{a}_j \right]$$ 

\section{\ds}
\label{sec:data}

\cut{
\begin{algorithm}[t]
\caption{Annotation process} \label{alg:linking}
\begin{algorithmic}[1]
\State detect \textbf{initial entity annotations} using article links, named entity linking, and coreference.
\State populate the \textbf{local knowledge graph}:
    \Indent
    \For {each \textit{token} in the article}
        \State \textbf{link the entity} of the \textit{token}.
        \State \textbf{annotate relations} to prior entities.
        \State \textbf{expand} \textit{token} in the graph.
        \State \textbf{prune} stale entities.
    \EndFor
    \EndIndent
\end{algorithmic}
\end{algorithm}
}
\begin{table*}[t]
\centering
\tabcolsep=0.11cm
\footnotesize
\begin{tabular}{rcccccccccccccccc}
\toprule
\bf Tokens& $x_t$ & Super & Mario & Land & is & a & 1989 & side & - & scrolling & platform & video & game& developed \\
\cmidrule(lr){3-5}
\cmidrule(lr){8-8}
\cmidrule(lr){9-11}
\cmidrule(lr){12-14}
\bf Mention type& $t_t$ & \multicolumn{3}{c}{\texttt{new}} & $\emptyset$ & $\emptyset$ & \texttt{related} & \multicolumn{3}{c}{\texttt{new}} & \multicolumn{3}{c}{\texttt{related}}  & $\emptyset$ \\
\bf Entity Mentioned& $e_t$ & \multicolumn{3}{c}{\colorbox[HTML]{1b9e77}{\ent{SML}}} & $\emptyset$ & $\emptyset$ & {\texttt{04-21-1989}} & \multicolumn{3}{c}{\ent{SIDE\_SCROLL}} & \multicolumn{3}{c}{\ent{PVG}}  & $\emptyset$ \\
\bf Relation& $r_t$ & \multicolumn{3}{c}{$\emptyset$} & $\emptyset$ & $\emptyset$ & \ent{pub date} & \multicolumn{3}{c}{$\emptyset$} & \multicolumn{3}{c}{\ent{genre}}  & $\emptyset$ \\
\bf Parent Entity& $p_t$ & \multicolumn{3}{c}{$\emptyset$} & $\emptyset$ & $\emptyset$ & \colorbox[HTML]{1b9e77}{\ent{SML}} & \multicolumn{3}{c}{$\emptyset$} & \multicolumn{3}{c}{\colorbox[HTML]{1b9e77}{\ent{SML}}}  & $\emptyset$ \\

\end{tabular}
\vskip 3mm
\begin{tabular}{ccccccccccccccccc}
$x_t$ & and & published & by & Nintendo & as & a & launch & title & for & their& Game & Boy & handheld & game & console & . \\
\cmidrule(lr){5-5}
\cmidrule(lr){8-9}
\cmidrule(lr){12-13}
\cmidrule(lr){14-16}
$t_t$ & $\emptyset$ & $\emptyset$ & $\emptyset$ & \texttt{related} & $\emptyset$ & $\emptyset$ & \multicolumn{2}{c}{\texttt{new}} & $\emptyset$ & $\emptyset$ & \multicolumn{2}{c}{\texttt{related}} & \multicolumn{3}{c}{\texttt{related}} & $\emptyset$\\
$e_t$ & $\emptyset$ & $\emptyset$ & $\emptyset$ & \colorbox[HTML]{e6ab02}{\ent{NIN}} & $\emptyset$ & $\emptyset$ & \multicolumn{2}{c}{\ent{LT}} & $\emptyset$ & $\emptyset$  & \multicolumn{2}{c}{\colorbox[HTML]{a6761d}{\ent{GAME\_BOY}}} & \multicolumn{3}{c}{\ent{HGC}} & $\emptyset$\\
$r_t$ & $\emptyset$  & $\emptyset$ & $\emptyset$ & \ent{pub} & $\emptyset$ & $\emptyset$ & \multicolumn{2}{c}{$\emptyset$} & $\emptyset$ & $\emptyset$ & \multicolumn{2}{c}{\ent{R:manu / platform}} & \multicolumn{3}{c}{\ent{instance of}} & $\emptyset$ \\
$p_t$ & $\emptyset$  & $\emptyset$ & $\emptyset$& \colorbox[HTML]{1b9e77}{\ent{SML}} & $\emptyset$ & $\emptyset$ & \multicolumn{2}{c}{$\emptyset$} & $\emptyset$ & $\emptyset$ & \multicolumn{2}{c}{\colorbox[HTML]{e6ab02}{\ent{NIN}} / \colorbox[HTML]{1b9e77}{\ent{SML}}} & \multicolumn{3}{c}{\colorbox[HTML]{a6761d}{\ent{GAME\_BOY}}} & $\emptyset$ \\
\bottomrule
\end{tabular}
\caption{\textbf{Example Annotation} of the sentence from Figure~\ref{fig:data}, including corresponding variables from Figure~\ref{fig:generative}. Note that \textit{Game Boy} has multiple parent and relation annotations, as the platform for \ent{Super Mario Land} and as manufactured by \ent{Nintendo}. Wikidata identifiers are made human-readable (e.g., \ent{SML} is Q647249) for clarity.}
\label{tab:example-annotation}
\end{table*}

Modeling aside, one of the primary barriers to incorporating factual knowledge into language models is that training data is hard to obtain.
Standard language modeling corpora consist only of text, and thus are unable to describe which entities or facts each token is referring to.
In contrast, while relation extraction datasets link text to a knowledge graph, the text is made up of disjoint sentences that do not provide sufficient context to train a powerful language model.
Our goals are much more aligned to the \emph{data-to-text} task~\cite{ahn2016neural,lebret2016neural,wiseman2017challenges,yang2017reference,gardent2017webnlg,ferreira2018enriching}, where a small table-sized KB is provided to generate a short piece of text; we are interested in language models that dynamically decide the facts to incorporate from the knowledge graph, guided by the discourse.

For these reasons we introduce the \ds{} dataset, consisting of (approximately) the same articles appearing in the \textit{WikiText-2} language modeling corpus,
but linked to the Wikidata~\cite{wikidata} knowledge graph.
Because the text closely matches, models trained on \ds{} can be compared to models trained on \textit{WikiText-2}.
Furthermore, because many of the facts in Wikidata are derived from Wikipedia articles, the knowledge graph has a good coverage of facts expressed in the text.
The dataset is available for download at: \url{https://rloganiv.github.io/linked-wikitext-2}.
Our system annotates one document at a time
, and consists of entity linking, relation annotations, and post-processing. 
The following paragraphs describe each step in detail.


\para{Initial entity annotations}
We begin by identifying an initial set of entity mentions within the text.
The primary source of these mentions is the human-provided links between Wikipedia articles.
Whenever a span of text is linked to another Wikipedia article, we associate its corresponding Wikidata entity with the span.
While article links provide a large number of gold entity annotations, they are insufficient for capturing all of the mentions in the article since entities are only linked the first time they occur.
Accordingly, we use the neural-el~\cite{GuptaSiRo17} entity linker to identify additional links to Wikidata, and identify coreferences using Stanford CoreNLP\footnote{\url{https://stanfordnlp.github.io/CoreNLP/}} to cover pronouns, nominals, and other tokens missed by the linker.

\para{Local knowledge graph} The next step iteratively creates a generative story for the entities using relations in the knowledge graph as well as identifies new entities.
To do this, we process the text token by token.
Each time an entity is encountered, we add all of the related entities in Wikidata as candidates for matching.
If one of these related entities is seen later in the document, we identify the entity as a parent for the later entity.
Since multiple relations may appear as \emph{explanations} for each token, we allow a token to have multiple facts.

\para{Expanding the annotations}
Since there may be entities that were missed in the initial set, as well as non-entity tokens of interest such as dates and quantities we further expand the entity annotations using string matching.
For entities, we match the set of aliases provided in Wikidata.
For dates, we create an exhaustive list of all of the possible ways of expressing the date (e.g. "\textit{December 7, 1941}", "\textit{7-12-1941}", "\textit{1941}", ...).
We perform a similar approach for quantities, using the \texttt{pint} library in Python to handle the different ways of expressing units (e.g. "\textit{g}", "\textit{gram}", ...).
Since there are many ways to express a numerical quantity, we only render the quantity at the level of precision supplied by Wikidata, and do not perform unit conversions.

\para{Example Annotation}
An example annotation is provided in Table~\ref{tab:example-annotation} corresponding to the instance in Figure~\ref{fig:data}, along with the variables that correspond to the generative process of the knowledge graph language model (KGLM).
The entity mentioned for most tokens here are human-provided links, apart from \tok{1989} that is linked to \ent{04-21-1989} by the string matching process.
The annotations indicate which of the entities are \emph{new} and \emph{related} based on whether they are reachable by entities linked so far, clearly making a mistake for \ent{side-scrolling game} and \ent{platform video game} due to missing links in Wikidata.
Finally, multiple plausible reasons for \ent{Game Boy} are included: it's the platform for \ent{Super Mario Land} and it is manufactured by \ent{Nintendo}, even though only the former is more relevant here.
Even with these omissions and mistakes, it is clear that the annotations are rich and detailed, with a high coverage, and thus should prove beneficial for training knowledge graph language models.

\begin{table}[tb]
    \centering
    \small
    \begin{tabular}{lrrr}
        \toprule
        &\bf Train & \bf Dev & \bf Test \\
        \midrule
        Documents       & 600       & 60        & 60  \\
        Tokens          & 2,019,195 & 207,982   & 236,062 \\
        Vocab. Size     & 33,558    &   -       &  - \\
        Mention Tokens  & 207,803   & 21,226    & 24,441 \\
        Mention Spans   & 122,983   & 12,214    & 15,007 \\
        Unique Entities & 41,058    & 5,415     & 5,625 \\
        Unique Relations & 1,291    & 484       & 504 \\
        \bottomrule
    \end{tabular}
    \caption{\textbf{\ds{} Corpus Statistics.}}
    \label{tab:dataset-stats}
\end{table}

\para{Dataset Statistics} Statistics for \ds{} are provided in Table~\ref{tab:dataset-stats}.
In this corpus, more than $10\%$ of the tokens are considered entity tokens, i.e. they are generated as factual references to information in the knowledge graph.
Each entity is only mentioned a few times (less than $5$ on average, with a long tail), and with more than thousand different relations.
Thus it is clear that regular language models would not be able to generate factual text, and there is a need for language models to be able to refer to external sources of information. 

\para{Differences from \textit{WikiText-2}}
Although our dataset is designed to closely replicate \textit{WikiText-2}, there are some differences that prevent direct comparison.
Firstly, there are minor variations in text across articles due to edits between download dates.
Secondly, according to correspondence with~\citet{merity2016pointer}, \textit{WikiText-2} was collected by querying the Wikipedia Text API.
Because this API discards useful annotation information (e.g. article links), \ds{} instead was created by directly from the article HTML.

\section{Training and Inference for KGLM}
\label{sec:algs}

In this section, we describe the training and inference algorithm for KGLM.

\para{Pretrained KG Embeddings}
During evaluation, we may need to make predictions on entities and relations that have not been seen during training.
Accordingly, we use fixed entity and relations embeddings pre-trained using TransE~\cite{bordes2013translating} on Wikidata.
Given $(p, r, e)$,  we learn embeddings $\mathbf{v}_p$, $\mathbf{v}_r$ and $\mathbf{v}_e$ to minimize the distance:
$$ \delta(\mathbf{v}_p, \mathbf{v}_r, \mathbf{v}_e) = \left \lVert \mathbf{v}_p + \mathbf{v}_r - \mathbf{v}_e \right \rVert ^2 . $$
We use a max-margin loss to learn the embeddings:
$$ \mathcal{L} = \text{max} \left(0, \gamma + \delta \left(\mathbf{v}_p, \mathbf{v}_r, \mathbf{v}_e \right) - \delta \left(\mathbf{v}_p', \mathbf{v}_r, \mathbf{v}_e' \right) \right) $$
where $\gamma$ is the margin, and either $p'$ or $e'$ is a randomly chosen entity embedding.

\para{Training with \ds}
Although the generative process in KGLM involves many steps, training the model on \ds{} is straightforward.
Our loss objective is the negative log-likelihood of the training data: 
$$ \ell(\Theta) = \sum_t \log p(x_t, \mathcal{E}_t | x_{<t}, \mathcal{E}_{<t}; \Theta), $$
where $\Theta$ is the set of model parameters.
Note that if an annotation has multiple viable parents such as \ent{Game Boy} in~\ref{fig:data}, then we marginalize over all of the parents.
Since all random variables are observed, training can performed using off-the-shelf gradient-based optimizers.

\para{Inference}~\label{sec:inference}
While observing annotations makes the model easy to train, we do not assume that the model has access to annotations during evaluation.
Furthermore, as discussed in Section~\ref{sec:gen-kg}, the goal in language modelling is to measure the marginal probability $p(\mathbf{x}) = \sum_{\mathbf{\mathbfcal{E}}} p(\mathbf{x},\mathbf{\mathbfcal{E}}) $ not the joint probability.
However, this sum is intractable to compute due to the large combinatorial space of possible annotations.
We address this problem by approximating the marginal distribution using importance sampling. 
Given samples from a proposal distribution $q(\mathbf{\mathbfcal{E}} | \mathbf{x})$ the marginal distribution is:
\begin{align*}
    p(\mathbf{x}) &= \sum_{\mathbf{\mathbfcal{E}}} p\left(\mathbf{x},\mathbf{\mathbfcal{E}}\right) 
        = \sum_{\mathbfcal{E}} \frac{p\left(\mathbf{x},\mathbf{\mathbfcal{E}}\right)}{q\left(\mathbf{\mathbfcal{E}} | \mathbf{x}\right)} q\left(\mathbf{\mathbfcal{E}} | \mathbf{x}\right) \\
        &\approx  \frac{1}{N} \sum_{\mathbfcal{E} \sim q} \frac{p\left(\mathbf{x},\mathbfcal{E}\right)}{q\left(\mathbfcal{E} | \mathbf{x}\right)}
\end{align*}
This approach is used to evaluate models in~\citet{ji2017dynamic} and~\citet{dyer2016recurrent}.
Following~\citet{ji2017dynamic}, we compute $q\left(\mathbfcal{E} | \mathbf{x} \right)$ using a discriminative version of our model that predicts annotations for the current token instead of for the next token.

\section{Experiments}~\label{sec:experiments}

To evaluate the proposed language model, we first introduce the baselines, followed by an evaluation using perplexity of held-out corpus, accuracy on fact completion, and an illustration of how the model uses the knowledge graph.

\subsection{Evaluation Setup}


\para{Baseline Models}
We compare KGLM to the following baseline models:
\begin{itemize}[nosep,leftmargin=3mm]
    \item\textbf{AWD-LSTM}\camready{\footnote{\scriptsize \url{https://github.com/salesforce/awd-lstm-lm}}}~\cite{merity2017regularizing}: strong LSTM-based model used as the foundation of most state-of-the-art models on \textit{WikiText-2}.
    \item\textbf{\textsc{EntityNLM}}~\cite{ji2017dynamic}: an LSTM-based language model with the ability to track entity mentions. Embeddings for entities are created dynamically, and are not informed by any external sources of information.
    \item\textbf{EntityCopyNet:} a variant of the KGLM where $t_t=\texttt{new}$ for all mentions, i.e. entities are selected from $\mathcal{E}$ and entity aliases are copied, but relations in the knowledge graph are unused.\\
\end{itemize}

\para{Hyperparameters}
We pre-train 256 dimensional entity and relation embeddings for all entities within two hops of the set of entities that occur in \ds{} using TransE with margin $\gamma=1$.
Weights are tied between all date embeddings and between all quantity embeddings to save memory.
Following~\citet{merity2017regularizing} we use 400 dimensional word embeddings and a 3 layer LSTM with hidden dimension 1150 to encode tokens.
We also employ the same regularization strategy (DropConnect~\cite{wan2013regularization} + Dropout\cite{srivastava2014dropout}) and weight tying approach.
However, we perform optimization using Adam~\cite{kingma2014adam} with learning rate 1e-3 instead of NT-ASGD, having found that it is more stable.

\subsection{Results}

\para{Perplexity}
We evaluate our model using the standard \textit{perplexity} metric: $\exp\left(\frac{1}{T}\sum_{t=1}^{T} \log p(x_t) \right)$.
However, perplexity suffers from the issue that it overestimates the probability of out-of-vocabulary tokens when they are mapped to a single UNK token.
This is problematic for comparing the performance of the KGLM to traditional language models on \ds{} since there are a large number of rare entities whose alias tokens are out-of-vocabulary.
That is, even if the KGLM identifies the correct entity and copies the correct alias token with high probability, other models can attain better perplexity by assigning a higher probability to UNK.
Accordingly, we also measure \textit{unknown penalized perplexity} (UPP) (a.k.a \textit{adjusted perplexity}) introduced by~\citet{UEBERLA1994153}, and used recently by \citet{ahn2016neural} and \citet{spithourakis2018numeracy}.
This metric penalizes the probability of UNK tokens by evenly dividing their probability mass over $\mathcal{U}$, the set of tokens that get mapped to UNK .
We can be compute UPP by replacing $p(\text{UNK})$ in the perplexity above by $ \frac{1}{\left|\mathcal{U} \right|} p(\text{UNK})$, where $\left|\mathcal{U} \right|$ is estimated from the data.

\begin{table}[tb]
    \centering
    \small
    \begin{tabular}{lcc}
        \toprule
            & \textbf{PPL}  & \textbf{UPP} \\
        \midrule        
        \textsc{EntityNLM}\textsuperscript{*}~\cite{ji2017dynamic} & 85.4 & 189.2 \\
        EntityCopyNet\textsuperscript{*} & 76.1 & 144.0\\
        AWD-LSTM\textsuperscript~\cite{merity2017regularizing} &  74.8 & 165.8 \\
        KGLM\textsuperscript{*} & \bf 44.1 & \bf 88.5 \\
        \bottomrule
    \end{tabular}
    \caption{\textbf{Perplexity Results} on~\ds. Results for models marked with * are obtained using importance sampling.
    }
    \label{tab:ppl}
\end{table}

We present the model perplexities in Table~\ref{tab:ppl}.
To marginalize over annotations, perplexities for the \textsc{EntityNLM}, EntityCopyNet, and KGLM are estimated using the importance sampling approach described in Section~\ref{sec:algs}.
We observe that the KGLM attains substantially lower perplexity than the other entity-based language models (44.1 vs. 76.1/85.4), providing strong evidence that leveraging knowledge graphs is crucial for accurate language modeling.
Furthermore, KGLM significantly outperforms all models in unknown penalized perplexity, demonstrating its ability to generate rare tokens.



\begin{table}[tb]
    \centering
    \setlength\tabcolsep{4pt}
    \footnotesize
    \begin{tabular}{lcccc}
        \toprule
            & \multirow{2}{*}{\bf \begin{tabular}{@{}l@{}}AWD-\\ LSTM\end{tabular}} & \multirow{2}{*}{\bf GPT-2} & \multicolumn{2}{c}{\bf KGLM} \\
            &  &  & Oracle & NEL \\
        \midrule        
        nation-capital &  0 / 0 &  \textbf{6} / \textbf{7} & 0 / 0 & 0 / 4 \\
        birthloc & 0 / 9 & 14 / 14 & \textbf{94} / \textbf{95} & 85 / 92 \\
        birthdate & 0 / 25 & 8 / 9 & \textbf{65} / \textbf{68} & 61 / 67 \\
        spouse &  0 / 0 & \textbf{2} / 3 & \textbf{2} / 2 & 1 / \textbf{19} \\
        city-state & 0 / 13 & \textbf{62} / \textbf{62} & 9 / 59 & 4 / 59 \\
        book-author & 0 / 2 & 0 / 0 & \textbf{61} / \textbf{62} & 25 / 28 \\
        \addlinespace
        \bf Average & 0.0/8.2 & 15.3/15.8 & \textbf{38.5}/\textbf{47.7} & 29.3/44.8 \\
        \bottomrule
    \end{tabular}
    \caption{\textbf{Fact Completion}. Top-$k$ accuracy (@1/@5,\%) for predicting the next token for an incomplete factual sentence. See examples in Table~\ref{tab:completion-examples}.}
    \label{tab:completion}
\end{table}

\begin{table*}[tb]
    \small
    \centering
    \vskip -1mm
\begin{tabular}{llcll}
  \toprule 
  &\bf Input Sentence & \bf Gold & \bf GPT-2 & \bf KGLM \\ 
  \midrule 
\multirow{2}{*}{\bf Both correct} &
Paris Hilton was born in \underline{\hspace{5mm}}	&\ent{New York City}&New& 1981\\
&
Arnold Schwarzenegger was born on \underline{\hspace{5mm}} &\ent{1947-07-30} & July & 30 \\

\addlinespace
\multirow{3}{*}{\bf KGLM correct} &
Bob Dylan was born in \underline{\hspace{5mm}}&\ent{Duluth}&	New&	Duluth\\
&
Barack Obama was born on \underline{\hspace{5mm}}&\ent{1961-08-04}&	January&	August \\
&
Ulysses is a book that was written by \underline{\hspace{5mm}}&\ent{James Joyce}&	a&	James\\

\addlinespace
\multirow{3}{*}{\bf GPTv2 correct} &
St. Louis is a city in the state of  \underline{\hspace{5mm}}&\ent{Missouri}		&	Missouri & Oldham  \\
&
Richard Nixon was born on \underline{\hspace{5mm}}&	\ent{1913-01-09} & January & 20 \\
&
Kanye West is married to \underline{\hspace{5mm}} &\ent{Kim Kardashian}&Kim&the\\

\addlinespace
\multirow{2}{*}{\bf Both incorrect} &
The capital of India is	 \underline{\hspace{5mm}}&\ent { New Delhi}	& the & a \\
&
Madonna is married to \underline{\hspace{5mm}}&	\ent{Carlos Leon} &	a &	Alex\\

  \bottomrule 
\end{tabular}

    \caption{\textbf{Completion Examples}. Examples of fact completion by KGLM and GPT-2, which has been trained on a much larger corpus. GPT-2 tends to produce very common and general tokens, such as one of a few popular cities to follow \tok{born in}. KGLM sometimes makes mistakes in linking to the appropriate fact in the KG, however, the generated facts are more specific and contain rare tokens. We omit AWD-LSTM from this figure as it rarely produced tokens apart from the generic \tok{the} or \tok{a}, or \tok{$\langle$UNK$\rangle$}. }
    \label{tab:completion-examples}
\end{table*}

\para{Fact Completion}
Since factual text generation is our primary objective, we evaluate the ability of language models to complete sentences with factual information. 
We additionally compare with the \emph{small} GPT-2~\cite{radford2019language}, a language model trained on a much larger corpus of text.
We select $6$ popular relations from Freebase, and write a simple \emph{completion} template for each, such as \tok{$X$ was born in \underline{\hspace{5mm}}} for the \rel{birthplace} relation.
We generate sentences for these templates for a number of $(X,Y)$ pairs for which the relation holds, and manually examine the first token generated by each language model to determine whether it is correct.

Table~\ref{tab:completion} presents performance of each language model on the relations.  The \emph{oracle} KGLM is given the correct entity annotation for $X$, while the \emph{NEL} KGLM uses the discriminative model used for importance sampling combined with the NEL entity linker to produce an entity annotation for $X$.

Amongst models trained on the same data, both KGLM variants significantly outperform AWD-LSTM; they produce accurate facts, while AWD-LSTM produced generic, common words.
KGLMs are also competitive with models trained on orders of magnitude more data, producing factual completions that require specific knowledge, such as birthplaces, dates, and authors. However, they do not capture facts or relations that frequently appear in large corpora, like the cities within states.\footnote{This is not a failure of the KG, but of the model's ability to pick the correct relation from the KG given the prompt.}
It is encouraging to see that the KGLM with automatic linking performs comparably to oracle linking.

We provide examples in Table~\ref{tab:completion-examples} to highlight qualitative differences between KGLM, trained on $600$ documents, and the recent state-of-the-art language model, GPT-2, trained on the WebText corpus with over $8$ million documents~\cite{radford2019language}.
For examples that both models get factually correct or incorrect, the generated tokens by KGLM are often much more specific, as opposed to selection of more popular/generic tokens (GPT-2 often predicts ``New York'' as the birthplace, even for popular entities).
KGLM, in particular, gets factual statements correct when the head or tail entities are rare, while GPT-2 can only complete facts for more-popular entities while using more-generic tokens (such as \tok{January} instead of \tok{20}).

\cut{
\begin{table}[tb]
    \centering
    \begin{tabular}{l}
    \toprule
    \multicolumn{1}{c}{\textbf{Input:} Barack Obama was born on \underline{\hspace{5mm}}}\\
    \addlinespace
    \small Fact:\hfill \ent{Barack Obama}, \rel{birthDate}, \ent{1961-08-04} \\
    \small Top-3 tokens:\hfill \tok{August}, \tok{4}, \tok{1961}\\
    \addlinespace
    \small Fact:~~ \ent{Barack Obama}, \rel{birthDate}, \colorbox{red!20}{\ent{2013-03-21}} \\
    \small Top-3 tokens:\hfill \tok{March}, \tok{21}, \tok{2013}\\
    \bottomrule
    \end{tabular}
    \caption{\textbf{Editing Facts}: Changing a fact in the knowledge graph results in a direct change to the output.}
    \label{tab:kg-edit}
\end{table}
}

\para{Effect of changing the KG}
For most language models, it is difficult to control their generation since \emph{factual} knowledge is entangled with generation capabilities of the model.
For KGLM, an additional benefit of its use of an external source of knowledge is that KGLM is directly controllable via modifications to the KG.
To illustrate this capability with a simple example, we create completion of \tok{Barack Obama was born on \underline{\hspace{3mm}}} with the original fact (\ent{Barack Obama}, \rel{birthDate}, \ent{1961-08-04}), resulting in the top three decoded tokens as \tok{August}, \tok{4}, \tok{1961}.
After changing the birth date to \ent{2013-03-21}, 
the top three decoded tokens become \tok{March}, \tok{21}, \tok{2013}.
Thus, changing the fact in the knowledge graph directly leads to a corresponding change in the model's prediction.

\section{Related Work}


\para{Knowledge-based language models}
Our work draws inspiration from two existing knowledge-based language models:

(i) \textsc{EntityNLM}~\cite{ji2017dynamic} which improves a language model's ability to track entities by jointly modeling named entity recognition and coreference.
Our model similarly tracks entities through a document, improving its ability to generate factual information by modeling entity linking and relation extraction.

(ii) The neural knowledge language model (NKLM)~\cite{ahn2016neural} which established the idea of leveraging knowledge graphs in neural language models.
The main differentiating factor between the KGLM and NKLM is that the KGLM operates on an entire knowledge graph and can be evaluated on text without additional conditioning information, whereas the NKLM operates on a relatively smaller set of predefined edges emanating from a single entity, and requires that entity be provided as conditioning information ahead of time.
This requirement precludes direct comparison between NKLM and the baselines in Section~\ref{sec:experiments}.

\para{Data-to-text generation}
Our work is also related to the task of neural data-to-text generation.
For a survey of early non-neural text generation methods we refer the reader to~\citet{reiter1997building}.
Recent neural methods have been applied to generating text from tables of sports statistics~\cite{wiseman2017challenges}, lists and tables~\cite{yang2017reference}, and Wikipedia info-boxes~\cite{lebret2016neural}.
The primary difference between these works and ours is our motivation. 
These works focus on generating coherent text within a narrow domain (e.g. sports, recipes, introductory sentences), and optimize metrics such as BLEU and METEOR score.
Our focus instead is to use a large source of structured knowledge to improve language model's ability to handle rare tokens and facts on a broad domain of topics, and our emphasis is on improving perplexity.


\para{General language modeling}
Also related are the recent papers proposing modifications to the AWD-LSTM that improve performance on \textit{Wikitext-2}~\cite{gong2018frage,yang2017breaking,krause2017dynamic}.
We chose to benchmark against AWD-LSTM since these contributions are orthogonal, and many of the techniques are compatible with the KGLM. 
KGLM improves upon AWD-LSTM, and we expect using KGLM in conjunction with these methods will yield further improvement. 

\section{Conclusions and Future Work}

By relying on memorization, existing language models are unable to generate factually correct text about real-world entities.
In particular, they are unable to capture the long tail of rare entities and word types like numbers and dates.
In this work, we proposed the \emph{knowledge graph language model} (KGLM), a neural language model that can access an external source of facts, encoded as a knowledge graph, in order to generate text.
Our implementation is available at:~\url{https://github.com/rloganiv/kglm-model}.
We also introduced \ds containing text that has been aligned to facts in the knowledge graph, allowing efficient training of the model.
\ds is freely available for download at:~\url{https://rloganiv.github.io/linked-wikitext-2}.
In our evaluation, we showed that by utilizing this graph, the proposed KGLM is able to generate higher-quality, factually correct text that includes mentions of rare entities and specific tokens like numbers and dates.

This work lays the groundwork for future research into knowledge-aware language modeling.
The limitations of the KGLM model, such as the need for marginalization during inference and reliance on annotated tokens, raise new research problems for advancing neural NLP models.
Our distantly supervised approach to dataset creation can be used with other knowledge graphs and other kinds of text as well, providing opportunities for accurate language modeling in new domains.

\section*{Acknowledgements}
First and foremost, we would like to thank Stephen Merity for sharing the materials used to collect the \textit{WikiText-2} dataset, and Nitish Gupta for modifying his entity linker to assist our work.
We would also like to thank Dheeru Dua and Anthony Chen for their thoughtful feedback.
This work was supported in part by Allen Institute of Artificial Intelligence (AI2), and in part by NSF award \#IIS-1817183.
The views expressed are those of the authors and do not reflect the official policy or position of the funding agencies.

\newpage
\bibliography{main}
\bibliographystyle{acl_natbib}


\end{document}